# Generalization is not a universal guarantee: Estimating similarity to training data with an ensemble out-of-distribution metric

W. Max Schreyer, BA[1, 2, 3], Christopher Anderson, MS[3], Reid F. Thompson, MD, PhD[1, 2, 3, 4, 5, 6, †]

1. Computational Biology Program, Oregon Health & Science University, Portland, Oregon, USA
2. Department of Biomedical Engineering, Oregon Health & Science University, Portland, Oregon, USA
3. Portland VA Research Foundation, Portland, Oregon, USA
4. Department of Radiation Medicine, Oregon Health & Science University, Portland, Oregon, USA
5. Department of Medical Informatics and Clinical Epidemiology, Oregon Health & Science University, Portland, Oregon, USA
6. Division of Hospital and Specialty Medicine, VA Portland Healthcare System, Portland, Oregon, USA

[†] Corresponding author: thompsre@ohsu.edu

## Abstract

Failure of machine learning models to generalize to new data is a core problem limiting the reliability of AI systems, partly due to the lack of simple and robust methods for comparing new data to the original training dataset. We propose a standardized approach for assessing data similarity in a model-agnostic manner by constructing a supervised autoencoder for generalizability estimation (SAGE). We compare points in a low-dimensional embedded latent space, defining empirical probability measures for *k*-Nearest Neighbors (kNN) distance, reconstruction of inputs and task-based performance. As proof of concept for classification tasks, we use MNIST and CIFAR-10 to demonstrate how an ensemble output probability score can separate deformed images from a mixture of typical test examples, and how this SAGE score is robust to transformations of increasing severity. As further proof of concept, we extend this approach to a regression task using non-imaging data (UCI Abalone). In all cases, we show that out-of-the-box model performance increases after SAGE score filtering, even when applied to data from the model's own training and test datasets. Our out-of-distribution scoring method can be introduced during several steps of model construction and assessment, leading to future improvements in responsible deep learning implementation.

## Background

The presence of generalization gaps, where machine learning performance degrades when a trained model encounters previously-unseen data, represents a critical ongoing challenge in the implementation of AI systems.[1,2] Model performance may suffer when the underlying distributions of input features for new data shift away from those learned during the training process. A baseline method for monitoring predictive uncertainty in neural networks without retraining is the maximum softmax prediction probability[3], where the highest output node value may decrease for out-of-distribution data points. While this technique has been improved with calibration of the softmax probabilities via temperature scaling[4,5], the approach has proved unreliable with increasingly-deformed input features and can be erroneously overconfident when predicting on unrecognizable images.[6,7] Ensemble methods have been proposed to improve the reliability of uncertainty measures, but this requires the simultaneous training of $m$ networks instead of a single model, increasing computational overhead.[8]

Whereas neural network prediction confidence is a black box measure of data similarity, there exist simple-to-understand visualization methods such as UMAP[9] and t-SNE[10], allowing users to examine similarity of points in high-dimensional space by localization patterns in 2 or 3D. While the resulting plots are appealing and easy to digest, the local and global structure of the data can become distorted by these methods of compression, limiting the effective use of distances as an "all-in-one" measure of data similarity.[11] Furthermore, these dimensionality-reduction techniques are not reproducible without degrading algorithmic performance (i.e. no multi-threading) or perpetuating the random initialization state.[12] Rabanser et al. introduce a method for quantifying dataset differences via dimensionality-reduction by embedding both a reference and novel dataset before statistically comparing the resulting distributions.[13] Other recent approaches use deep embeddings to calculate a latent distance metric for identifying out-of-distribution data, a key advantage of which is the ability to discriminate individual samples instead of reporting whole dataset statistical differences.[14,15]

The tradeoffs between explainability, quantifiability and robustness have thus far been barriers to a consensus approach to determining which individual samples are appropriate to use for a given machine learning model. We therefore propose the use of supervised autoencoders for generalization estimates (SAGE) as a universal approach to uncertainty estimation that draws from the strengths of previously-described methods.[16] SAGE scoring is introduced as a dataset companion which allows for the uncoupling of uncertainty estimation from downstream prediction tasks with separate, more complex models. We calculate a combined out-of-distribution score using three model-intrinsic measures of uncertainty estimation and show examples of outlier detection for classification tasks using MNIST and CIFAR-10 and a regression task using the UCI Abalone dataset. Finally, we show how filtering outliers using the combined out-of-distribution score improves generalization to separate, stronger classification and regression models, even with perturbed and corrupted data.

## Results

We demonstrate a supervised autoencoder (SAE) framework for faithfully encoding MNIST training data images in two dimensions (Fig. 1A), with low error in image reconstruction across held-out test images (∆ train reconstruction error = 0.005, $n$ = 8,000) (Fig 1B). Designed in part to capture digit identity in the latent space through multitask learning, the model also demonstrates excellent classification performance on held-out test images (f1 = 97.9) (Fig. 1C). Moreover, the latent space distribution of MNIST training data is closely approximated by encoded MNIST test data, which is drawn from the same original image distribution. (Fig. 1D). Note that we observed local differences in data density across the embedded latent space, along with local differences in average reconstruction error and calibrated classification confidence (Supplementary Fig. 1). Importantly, modified test images (i.e. transformed data that intentionally deviate from native MNIST examples) mapped to different areas of the latent space, with increasing severity of transform mapping to lower density, lower confidence regions of the latent space (Fig. 1E). Indeed, each transformation of test images results in measurable deviations from the original latent space encoding, with increased distance to the $k$-nearest training points (Fig. 1F), increased image reconstruction error, and decreased calibrated classifier confidence in nearly all cases (Supplementary Fig. 2). We found that minimally transformed images (e.g. "low" elastic deformation) tend to map closely to the original image set, whereas larger deformations (e.g. "high" elastic deformation) are significantly more distinct in their latent space embedding (Fig. 1E, 1F, Supplementary Fig 2). We identified a small minority of so-called "imposter" transformations, where vertical or horizontal geometric transforms resulted in effective misclassification (e.g. vertically-flipping a '5' will be read as '2') (Fig. 1E, Supplementary Fig. 3) and removed such imposters from subsequent analysis.

Recognizing that latent space density (as assessed by $k$NN distance), reconstruction error, and classifier confidence are all distinct and independent phenomena (Supplementary Fig. 4), we created an ensemble score (supervised autoencoder for generalizability estimation [SAGE] score) based on the exceedance probability of training data with respect to each of these measures (Fig. 2A, Methods). This ensemble approach clearly separated original MNIST train and test sets from transformed data (mean SAGE scores for train = 0.444, test = 0.441 and transformed = 0.075) (Fig. 2B). We found that the lowest SAGE scores identify outlier images among MNIST training and testing data (Fig. 2C-D), and are particularly discriminative for transformed images, where the majority of scores were zero or near-zero. More mild image transforms of MNIST, such as "low" elastic deformation, are prominent only towards the 90th percentile of SAGE scores for the transformed test dataset, where scores begin to increase appreciably (Fig. 2E). The lowest probability scores were associated with high degrees of pixel intensity changes including pixel inversions and heavy Gaussian noising.

In an attempt to improve out-of-the-box performance of an independent ResNet18 model trained on MNIST, we sought to leverage SAGE score as a data filter to ensure similarity of input data to the model's own training data. We first demonstrate that our combined score succeeds as a tunable filter selectively distinguishing outliers (transformed data) while preserving original training and testing data (Fig. 3A-C), with filter threshold values corresponding inversely to anticipated original dataset retention (e.g. threshold of 0.1 retains 93.7% of the train dataset).

We next note that the independent ResNet18 model performs exceedingly well on MNIST held-out test data (f1 = 0.99) but shows degraded performance on transformed data as expected (f1 = 0.76); whereas, transformed dataset performance improves significantly with even mild SAGE score filtering (e.g. f1 = 0.90 for threshold of 0.05) (Fig. 3D-F). We demonstrated that the SAGE score outperforms thresholding on the exceedance probabilities of individual model components, leading to higher separate model performance (Supplementary Fig. 5). We also note that out-of-the-box model performance can be improved even for data used during the training process, with increasing filter stringency improving observed accuracy (Supplementary Fig. 6, A-C).

Given the relative simplicity of MNIST as a use case, we next sought to apply our approach to a more complex image classification task using the CIFAR-10 dataset, which contains three color-channel data (3,072 RGB pixel values per image). We applied a panel of image perturbations of different intensities to the original CIFAR-10 test dataset (Fig. 4A), and demonstrate that a 2-dimensional latent space embedding, in this case using a deeper architecture and introducing a contrastive loss term, can faithfully encode distinct image clusters in this dataset (Fig. 4B). Reasoning that increasing dimensionality of the latent space could further improve performance, we demonstrated a reduction in overall training loss up to 16 dimensions, after which performance plateaued (Supplementary Fig. 7). Applying this 16-dimensional SAE to the train, test and transformed test CIFAR-10 image sets, we reproduce our findings from MNIST, where increasing SAGE score identifies increasing severity of image transformation, while minimal transformations (e.g. horizontal flip) behave similarly to untransformed test data (Fig. 4C-E). Importantly, SAGE score-based filtering improved performance of a separate out-of-the-box ResNet34 model pre-trained on CIFAR-10, particularly when applied to transformed images (average precision (AP) of 0.86 with SAGE and 0.44 without SAGE, 0.2 threshold value (Fig. 4F-H).

Finally, we sought to explore the potential of this approach for regression tasks. For proof-of-principle, we fit an SAE model to the UCI Abalone dataset, compressing the input feature space down to a single latent dimension (Fig. 5A). As before, SAGE scores for most transformed data points revealed significantly lower values compared with the training data set distribution (Fig. 5B), and filtering based on score demonstrated favorable exclusion of transformed data with relative retention of training and testing data (Fig. 5C-E). Finally, we applied score-based filtering of input data to a separate random forest regression model trained on the original training dataset, demonstrating improved root mean squared error (RMSE) of predictions with increasing threshold values, including for samples within the original train and test sets (Fig. 5F-H, Supplementary Fig. 6, H-J).

## Discussion

We present here a flexible, model-agnostic, dataset-focused approach for prospective detection of out-of-distribution data points. We demonstrate the SAGE score's potential use as a selective filter of input data prior to model application, with several advantages over existing techniques. Moreover, SAGE is applicable to different datasets and tasks, including both classification and regression, and has potential implications for model development (e.g. via outlier identification and model refinement), refining or adapting existing models to new data, and supporting regulatory review and post-market surveillance. To our knowledge, this is the first approach that standardizes generalizability estimation across modalities and tasks while prioritizing interpretability and remaining sensitive to covariate shifts in the underlying data.

The SAGE score is an ensemble metric that combines three independent measures of out-of-distribution estimation, compensating for the relative weaknesses of each component. For example, the model encoder outputs latent embeddings which allow for visualization in a low-dimensional space but does not exclusively rely on compressed distances as a measure of similarity, an attribute that has been shown to be problematic for popular methods like t-SNE and UMAP. Data reconstructions, assessed by the L2 norm, can also yield misleading results as is the case when CIFAR-10 images are subjected to a Gaussian blur transform. While Gaussian blurring results in a lower overall reconstruction error than that of original train images, higher kNN distance to training points in the embedding space and lower classification confidence allow these images to be filtered out. Furthermore, unrecognizable images such as unnormalized Gaussian noise in RGB color channels exhibit perfect classifier confidence but are easily detected and removed by a combination of reconstruction error and kNN distance.

Despite these strengths, we note several limitations to our work. Our study focuses on classification and regression with image and biological data as our primary machine learning tasks, neglecting any number of other common problems and data modalities (e.g. histopathological image segmentation or time-series forecasting with economic or weather pattern data). We also concede the potential to further improve SAGE performance through increasing model size, complexity, and encoder pretraining, as well as alternative or additional architectures. For instance, the inclusion of Bayesian dropout for neural network classifiers could improve variational inference without the need for retraining pre-existing models.[17,18] Other approaches for Bayesian inference have been suggested for neural network regression, and could be similarly applied.[19]

Furthermore, we do not perform data augmentation before training and our method can therefore be considered a form of normative modeling.[20] Prior work by *Hendrycks et al.*[21] has shown that inclusion of few augmented examples during training can improve the robustness of subsequent classifier confidence measures to outliers, a simple method that negates the use of more expensive generative models to create synthetic data.[22] Upsampling training data that has a low similarity score to itself could further augment the training process and improve generalization in a complementary manner, however, these iterations are considered out-of-scope in this proof-of-principle study.

We further note that SAGE scoring is unable to distinguish "imposter" data examples (e.g. where a vertical flip of a '5' in MNIST may be mistakenly recognized as a '2'). We did not observe any such instances within our transformed test sets of CIFAR-10 and therefore expect this phenomenon to be rare in real-world applications as images increase in complexity. Importantly, we also note that SAGE may expose sensitive, private, and/or proprietary details about a model's training dataset through the retention of both encoder and decoder elements in addition to the full latent space embedding. We envision the possibility of privacy-preserving implementations of this work, but note that these are out-of-scope in the current study. Future work will focus on the extension of out-of-distribution estimation to a wider range of tasks and modalities, including more complex biomedical imaging datasets, and the inclusion of improved measures of intrinsic uncertainty.

## Methods

### Datasets

The MNIST[23] dataset was downloaded using the torchvision package (version 0.17.2). MNIST consists of 28 x 28 pixel grayscale images of handwritten digits (0 - 9) and comes pre-split into training ($n$ = 60,000) and testing ($n$ = 10,000) sets, with 6,000 and 1,000 images per class respectively. We randomly divided the test set into class-balanced, held-out test ($n$ = 8,000) and validation ($n$ = 2,000) sets in order to set aside images for classifier calibration.

The CIFAR-10[24] dataset was downloaded using torchvision and consists of 32 x 32 pixel RGB color images of ten vehicle and animal classes. Like MNIST, CIFAR-10 is pre-split into a training ($n$ = 50,000) and testing ($n$ = 10,000) set which we randomly subdivided further into held-out test ($n$ = 8,000) and validation ($n$ = 2,000) sets, ensuring class balance. The image classes consist of: 'Airplane', 'Automobile', 'Bird', 'Cat', 'Deer', 'Dog', 'Frog', 'Horse', 'Ship', and 'Truck'. Importantly, 'Automobile' and 'Truck' vehicle classes consist of only cars and tractor-trailers, respectively, in order to reduce label overlap, whereas 'Airplane' and 'Ship' consist of different grades of planes (e.g. commercial passenger jets, military jets) and watercraft (e.g. leisure boats, commercial shipping vessels). All animal classes include multiple species or breeds. CIFAR-10 images also exhibit a variety of naturally-occurring viewer perspectives and subject color patterns, lending to the increased complexity of this dataset.

The UCI Abalone dataset was downloaded from the UC Irvine Machine Learning Repository website (https://archive.ics.uci.edu/dataset/1/abalone) and is included in our project repository as a CSV file. The dataset was adapted from a 1994 technical report[25] and consists of 4,177 examples of 8 animal phenotypes and body measurements including Sex, Length, Diameter, Height, Whole Weight, Shucked Weight, Viscera Weight, and Shell Weight, with the number of inner-shell rings representing the ground-truth labels. We split examples into training (80%), held-out test (16%) and validation (4%) datasets.

### Data Transformations

We built and applied a panel of image transformations to MNIST and CIFAR-10 held-out test images using the v2 transform module of torchvision's library. The panel included a 100% horizontal flip, 100% vertical flip, 100% pixel value inversion, Gaussian blur (kernel size = 5, sigma = 2), Gaussian noise ("low", sigma = 0.2; "high", sigma = 0.8) and elastic stretching ("low", alpha = 50; "high", alpha = 200). For CIFAR-10 we included two additional photometric transformations: a 100% solarize filter (threshold = 0.75) and 100% posterize filter (bits = 2). All MNIST and CIFAR-10 images were converted to torch float32 data types, scaled and normalized using the following values before use in training and analysis: MNIST (mean = [0.1307], std dev = [0.3081]), CIFAR-10 (mean = [0.4914, 0.4822, 0.4465], std dev = [0.247, 0.243, 0.261]).

For the UCI Abalone dataset, we introduced custom transformations of the testing data including the random addition of Gaussian noise (Low σ = 0.05, High σ = 0.5), inverting features (1 - feature value), randomly dropping feature columns (Low $n$ = 1, High $n$ = 3) and multiplying the features by a factor of 2.0 or 0.5 to simulate abalone species with larger (factor of 2) or smaller (factor of 0.5) body proportions while keeping the number of rings constant. All features were standardized by removing the train set mean and scaling to unit variance before training and testing our model.

### Model Architectures

All supervised autoencoder (SAE) models consisted of a neural network encoder, a neural network decoder and third task-focused neural network module. All models were built using PyTorch (version 2.2.2) and python (version 3.10.14).

For MNIST, we constructed an encoder module with two convolutional layers (kernel size = 3, stride = 1, padding = 1) followed by 2D batch normalization and max pooling (kernel size = 2). The last two layers of our encoder were fully-connected from the flattened output of max pooling. The classifier module consisted of a two-layer fully-connected network using the encoder's latent embedding as its input, with 20 and 10 layers respectively. The decoder architecture for MNIST mirrored the encoder, with two fully-connected layers followed by unflattening and max un-pooling (kernel size = 2), after which two de-convolutional layers (kernel size = 3, stride = 1, padding = 1) return the original image size ([batch, 1, 28, 28]). All layers are followed by a Leaky RELU activation function and we use dropout (p = 0.2) between convolutional/de-convolutional layers.

For CIFAR-10, we instantiated a ResNet18 model from pytorch with default ImageNet pre-trained weights as the encoder module. We re-initialized the last fully-connected encoder layer before training. The classifier module consisted of two fully-connected layers using dropout (p = 0.2), with 20 and 10 layers respectively. The decoder contained a fully-connected layer with 1,024 nodes followed by unflattening and three de-convolutional layers (kernel size = 4, padding = 1, stride = 2). Like MNIST, we use Leaky RELU activation for all three modules.

The UCI Abalone model has four layer, fully-connected encoder and decoder modules, each followed by Leaky ReLU activation and dropout (p = 0.2) with 64, 32, 16 and 1 nodes respectively. The regressor module consists of three fully-connected layers with a single output node and no activation function using 32, 16, 8 and 1 node layers. All non-final layers of the encoder, decoder and regressor use Leaky RELU activation.

Full project code is available on GitHub (https://github.com/pdxgx/latent).

### Model Training

Training was performed on a laptop with a 6-core CPU and 32GB of RAM. For MNIST, we trained our supervised autoencoder model over 20 epochs with early stopping. We used an Adam optimizer with a learning rate of $3\times10^{-4}$ and batch size of 64. Decoder loss was measured using the mean squared error (MSE) loss function and classification loss was measured using cross-entropy loss. The total loss was calculated as the unweighted sum of the decoder and classifier loss terms. We utilized the pre-split MNIST training set ($n$ = 60,000) to fit the model without inclusion of any image transformations.

For CIFAR-10, we implemented a two-stage training process, each occurring over 10 epochs (20 epochs total) with a batch size of 64. The first stage only involved training the encoder and classifier weights with an Adam optimizer with a learning rate of $3\times10^{-4}$. We used cross entropy loss to quantify classification error and included a center loss term down-weighted by a coefficient (alpha = 0.1). We randomly-initialized a cluster center coordinate for each of the 10 classes. The first training phase maximized the distance between cluster centers, yielding improved latent separation of the image classes. For the second stage, we trained the encoder, decoder and classifier using a second Adam optimizer and learning rates of $1\times10^{-4}$, $3\times10^{-4}$ and $1\times10^{-5}$ respectively. We used different learning rates within the stage two optimizer to allow for the simultaneous training of the decoder and preservation of the latent embedding structure established during the first stage. The decoder and classifier loss terms were quantified using the MSE loss and cross entropy loss respectively. The total loss for stage two was calculated as the unweighted sum of decoder and classifier error.

Our UCI Abalone model was trained over 100 epochs with an Adam optimizer and a learning rate of $3\times10^{-4}$. We used MSE for both the decoder and regressor loss functions, and the total loss was the unweighted sum of these terms.

### Temperature Scaling

After the training process for MNIST and CIFAR-10, we calibrated the autoencoder classifier modules with temperature scaling. For each dataset, we classified all validation set images using the trained models and divided the raw logits by a tunable parameter, $T$, in order to align model predictions with the true likelihood of correct predictions. We used cross entropy loss and a L-BFGS optimizer (learning rate = 0.01, batch size = 64) to tune $T$ over one epoch for each model.

### *k*-Nearest Neighbors Distance

The training split for each dataset was designated as the 'reference' embedding for both classification and regression analyses. The reference data was compressed using the trained model encoder and a Balltree[26] was fit to the resulting latent space. Each test example underwent the same encoding process and the tree was queried using the latent coordinates to determine the average distance to the point's $k$-Nearest Neighbors (kNN). The value of $k$ was 100 for MNIST and CIFAR-10, and 20 for embedded Abalone data.

### SAGE Scoring

For a given training dataset, we used the corresponding trained supervised autoencoder model to generate the latent embedding coordinates, the reconstructed input and task-based prediction, either regression value or calibrated classifier confidence. We took the overall distributions of training set measures — log average kNN distance to embedded training points, reconstruction error as assessed by MSE loss and an uncertainty measure from the model's supervised component (e.g. negative log of classifier confidence) — and sorted them from low to high, determining the complementary cumulative density function (CCDF) for each output measure distribution. For each new data point, for each of the three output measures, we determined where the observed value falls within the training CCDFs, yielding the probability of finding a training value more extreme than the observed value. The SAGE score was calculated as the geometric mean of the three observed measure probabilities for each data point (Figure 2A). This enables the calculation of SAGE scores for any point, including for the training data points that make up the underlying CCDFs.

### Pre-trained ResNet Models & Precision Recall Curves

Pre-trained ResNet models for MNIST and CIFAR-10 were initialized using the timm[27] (version 1.0.12) library and incorporated into our workflow for assessing the effects of filtering data points based on SAGE score thresholds. We did not make any modifications and these models were used as out-of-the-box classifiers on our thresholded train, test and transformed datasets.

Precision recall curves were calculated using the scikit-learn[28] (version 1.4.2) PrecisionRecallDisplay command, and average precision values were calculated as the weighted mean of precisions achieved at each prediction threshold value, where the increase in recall from the previous threshold is used as the current weight.

### Random Forest Regression

The UCI Abalone training set was used to train a separate random forest regressor model from scikit-learn. We performed grid search cross-validation to determine the best model parameters, testing a variable number of estimators ([25, 50, 75, 100]), tree depths ([5, 10, 15, 20, 40]) and maximum features ([2, 4, 6, 8]). The best model had 50 estimators, a tree depth of 15 and used a maximum of 2 features. Regression error was assessed as the root mean square error (RMSE) between the number of inner-shell rings and predicted values for the train, test and transformed test sets.

### Score Thresholding & Performance Evaluation

SAGE scores were calculated for all examples in the MNIST, CIFAR-10 and UCI Abalone datasets as described above. For each set, data was filtered at six SAGE score values ([0.0, 0.01, 0.05, 0.1, 0.15, 0.2]) where samples greater than or equal to the threshold were retained and all others were discarded. Retained samples were passed to the separate, ResNet or random forest regression models and predictions were recorded. For MNIST and CIFAR-10 we used sci-kit learn's LabelBinarizer to one-hot encode labels and PrecisionRecallDisplay to create micro-averaged precision-recall curves from ResNet predictions. We repeated this process for the training, test and transformed test data separately, calculating average precision at each score threshold. Abalone predictions were assessed using sci-kit learn's root_mean_square_error function and visualized as matplotlib scatterplots.

## Disclaimer

The contents do not represent the views of the US Department of Veterans Affairs or the US Government. No terms were censored from this text.

# Figures

Figure 1

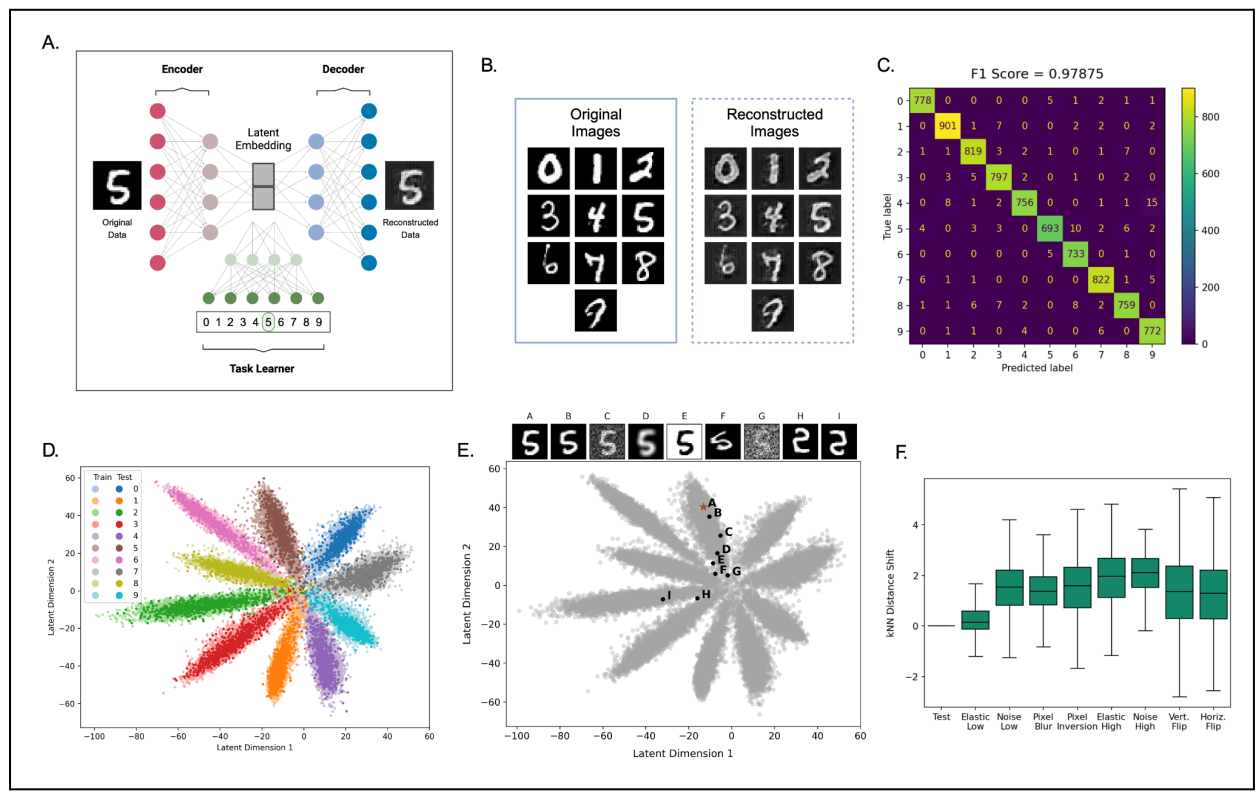

Overview of the design and outputs of a supervised autoencoder for generalization estimates (SAGE) model. A) Schematic of the SAGE architecture for MNIST data with three labeled components: an Encoder, a Decoder and a Task Learner (i.e. neural network digit identity classifier). Input image (Original Data) is shown at left, with a two-dimensional Latent Embedding shown in center, and a corresponding output image (Reconstructed Data) at right. B) A representative image from each class is shown at left, with corresponding reconstructed images at right. C) Confusion matrix of Task Learner module performance on MNIST test data, with each row corresponding to actual label (y-axis), each column corresponding to predicted label (x-axis), and each cell depicting the number of images having the indicated actual and predicted labels. Color scale is shown at right, and overall F1 score is displayed above. D) Scatter plot of MNIST training and testing images, with each point corresponding to a single image embedded in two latent dimensions whose values are shown on x- and y-axes, respectively, and whose actual class label is indicated by color according to the upper-left legend, with light and bold shading representing training and test images, respectively.. E) Two-dimensional embedding of MNIST training data is shown as a scatterplot in gray, with a representative test image (labeled "A") and corresponding transformed images shown above, with corresponding latent space embedding shown as labeled points "A" through "I" (A: Reference, B: Elastic Low, C: Noise Low, D: Gaussian Blur, E: Pixel Inversion, F: Elastic High, G: Noise High, H: Vertical Flip, I: Horizontal Flip). Reference image point is colored brown, corresponding to its class label in Figure 1D. F) Boxplots of different image transformations are shown and labeled along the x-axis, while the y-axis shows the change in the distance to the nearest *k* training neighbors in the embedded latent space relative to the untransformed test image's kNN distance. The black line across each box is the median kNN distance shift for that image set and the bottom and top edges of each box are the first and third quartile respectively. The whiskers extend past each box edge ± (1.5 x IQR).

## Figure 2

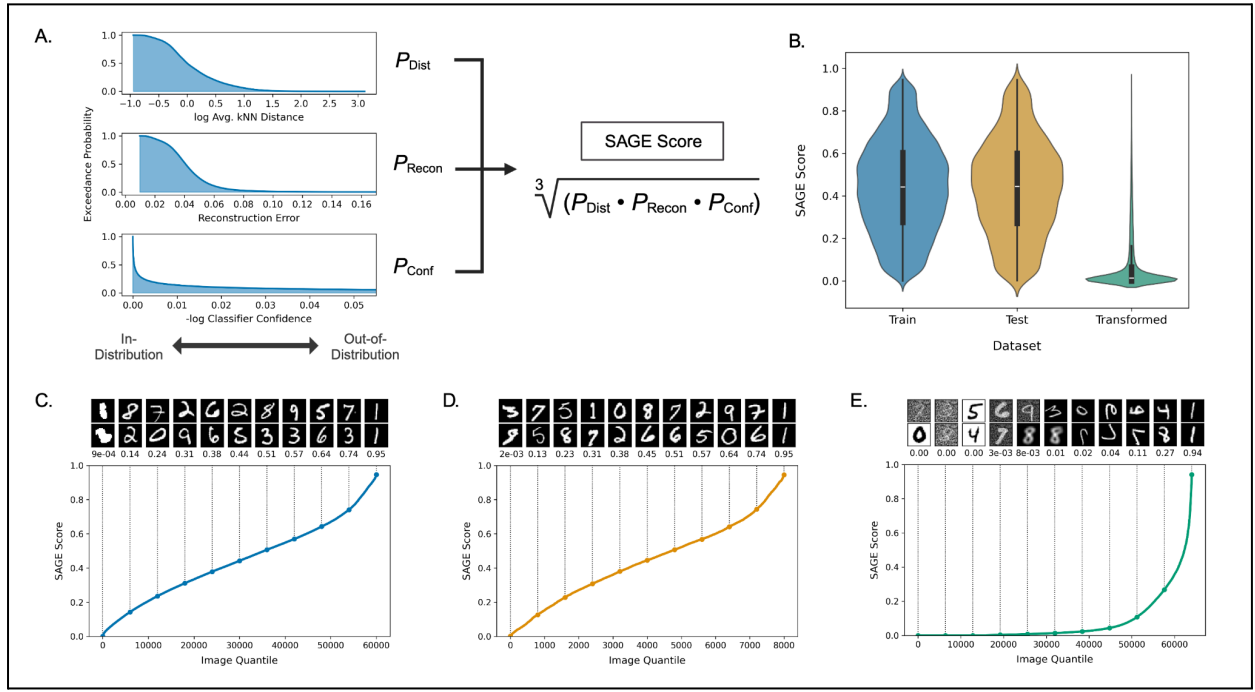

Overview of the SAGE scoring procedure and comparison of scores for the train, test and transformed test MNIST image sets. A) Each panel represents one of the three output measures of the trained supervised autoencoder model. For each measure, the complementary cumulative density curve shows the distribution of train set values, with log or negative log operations performed to ensure curves read from left (in-distribution) to right (out-of-distribution). Y-axis values represent the exceedance probability for any x-axis value, or the likelihood that a point at least as extreme as that measure is observed in the training data (Methods, SAGE Scoring). For any point, the SAGE score is calculated as the geometric mean of its three exceedance probability values. B) SAGE scoring separates transformed images from original train and test data. The y-axis shows the SAGE score calculated using the trained MNIST model. Each dataset is plotted along the x-axis in a different color, with all transformed images grouped together into one violinplot ($n$ = 64,000). The shape of each violin plot shows the concentration of SAGE score values, where the wider the plot the higher the number of images with that value. Violinplots also show embedded boxplots with the white line representing the median value for each dataset. C - D) Quantile plots for MNIST train ($n$ = 60,000), test ($n$ = 8,000) and transformed test ($n$ = 64,000, 8 transforms) sorted by SAGE score. Y-axis shows the calculated SAGE score with the x-axis counting the tally of the number of images belonging to each dataset. At each decile, a vertical dotted line leads to the quantile SAGE score value and the two nearest MNIST images.

Figure 3

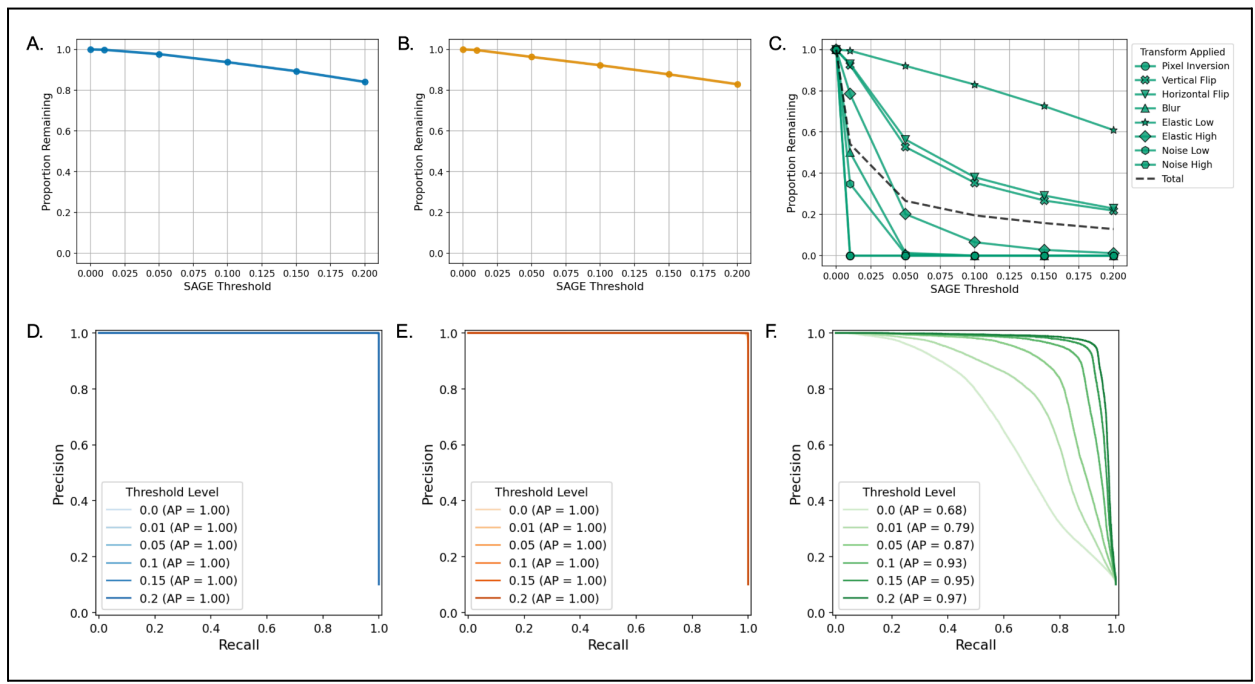

Effects of SAGE score filtering on performance of a separate ResNet MNIST classifier. A - B) Line plots for train and test datasets with the x-axis corresponding to SAGE score thresholds where images with score values lower than the threshold are removed. The plots show five threshold values: 0.0 (control), 0.01, 0.05, 0.1, 0.15, and 0.2. The remaining proportion of the original dataset is calculated and shown on the y-axis. C) For the transformed test dataset, each transform is represented as a distinct line with a different marker shape indicated in the upper-right legend, and the change in dataset proportion across all transforms in aggregate is shown as a black dashed line. D - F) The remaining images from each SAGE score threshold are used to predict classes with a separate ResNet18 classifier model trained on MNIST. Precision and recall of predictions are calculated across the range of classifier probability values, with the resulting points forming a precision recall (PR) curve. Curves are calculated for each SAGE score threshold level, with the color density increasing as the threshold values rise. Average precision (AP) values are used to summarize performance and are located in the plot legend alongside the threshold values.

Figure 4

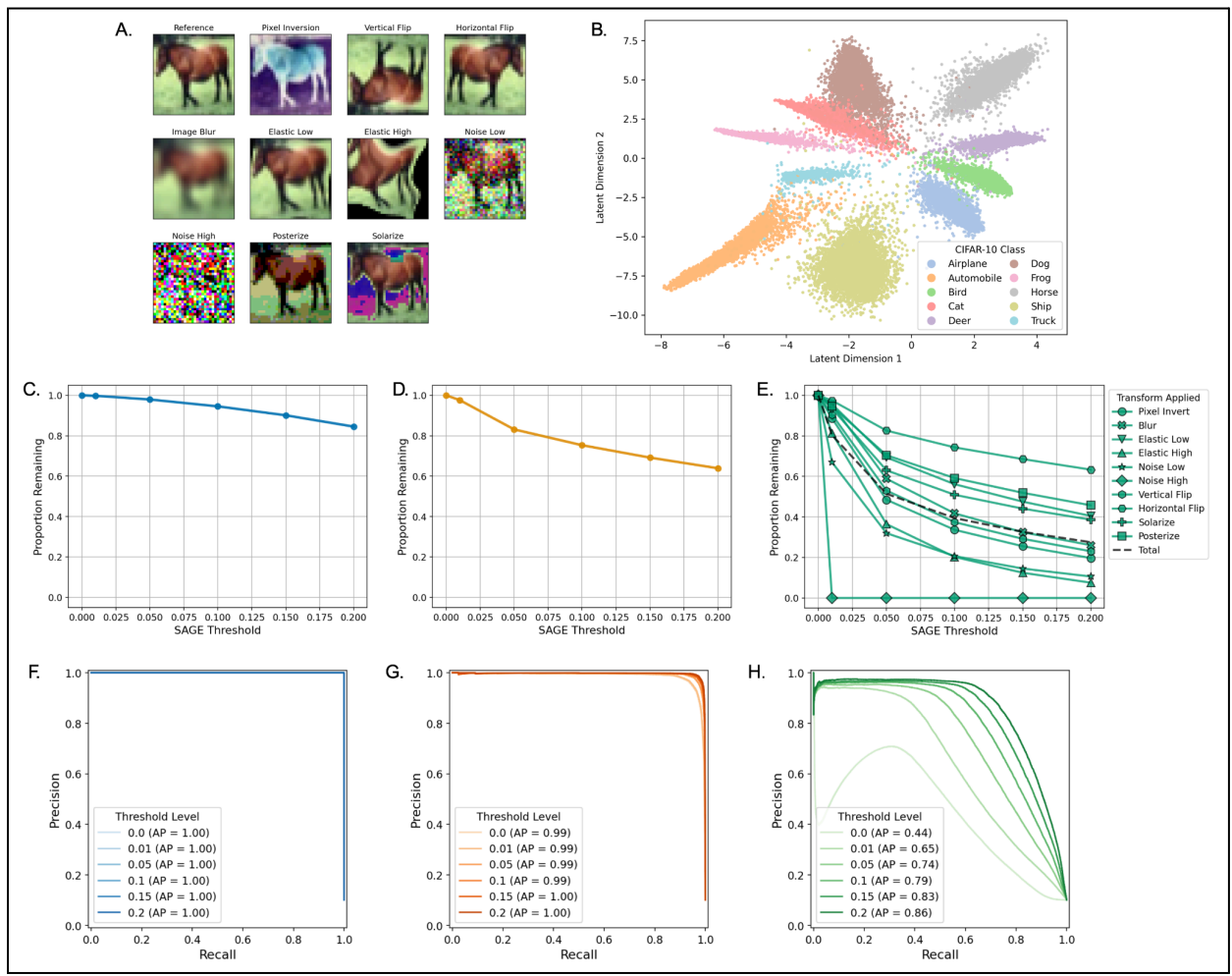

Training and evaluating SAGE on CIFAR-10 images with and without image transforms. A) The panel of transformations applied to the CIFAR-10 dataset, with both geometric and photometric deformations altering the appearance and RGB pixel values of the underlying image. We show representative image examples from a single class (Horse). B) CIFAR-10 train images shown as points on a 2D scatter plot after dimensionality-reduction with the trained encoder. Images were compressed from 3,072 pixels (originally [3, 32, 32] C x H x W) down to a 2D latent embedding using the ResNet encoder pre-trained on ImageNet and fine-tuned on the CIFAR-10 train image set. Each class is shown as a distinct color, with the descriptive name of the classes shown in the plot legend. C - D) SAGE scores calculated for all train and test CIFAR-10 images using a trained supervised autoencoder with a 16-dimensional latent embedding. Y-axis shows the proportion of the original datasets remaining after thresholding on x-axis SAGE score values. The same 5 threshold values as MNIST were used for analysis with CIFAR-10. E) Each of the 10 image transformations are given their own curve, with a unique marker symbol at each threshold value. An overall proportion curve is shown as a dashed black line. F - H) Precision recall curves for CIFAR-10 train, test and transformed images after SAGE thresholding and prediction with a pretrained ResNet34 model.

Figure 5

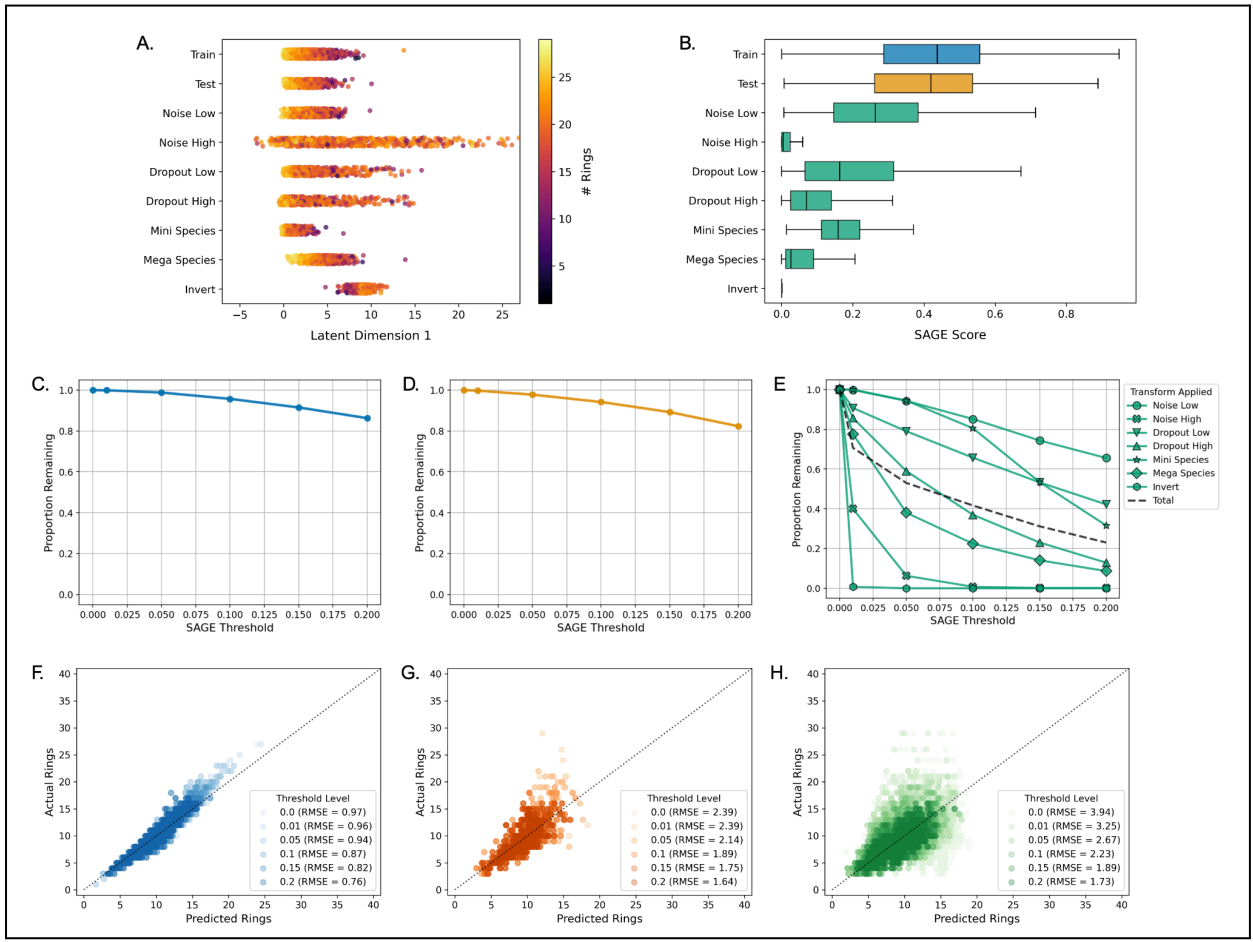

Analysis of a trained supervised autoencoder with a regressor module for predicting the inner-shell rings of *Haliotis rubra* from phenotype data. A) The encoder module compresses the 8 input features down to a single latent vector. Embedding values are shown as a scatterplot for train, test and transformed test sets, where the x-axis is the 1D latent coordinate and the y-axis is the embedded dataset. The number of ground-truth inner shell rings is depicted for each dataset as a color gradient of embedded points, with a larger number of rings corresponding to a lighter color yellow. B) SAGE scores were calculated using the trained model and represented as horizontal box plots. The x-axis shows the score values with the y-axis ticks showing the name of each of the datasets used in analysis. Box edges denote the first and third quartiles for the leftmost and rightmost edges respectively, with the median depicted as a black line. Boxplot whiskers extend ± (1.5 x IQR) beyond the box edges. Train, test and transformed test sets are shown in separate colors, with transforms all depicted in light green. C - D) Scores for train and test data at or below 5 x-axis threshold values were removed and shown as line plots, using the same thresholds as MNIST and CIFAR-10 datasets which are depicted as circular markers. Proportion of remaining examples is shown on the y-axis. E) Transformed test examples undergo the same SAGE score thresholding process, with each transform plotted as a line using a unique marker. The proportion of overall remaining transformed examples at each threshold level is plotted as a dashed black line. F - H) Data points passing each successive SAGE score threshold were used with a separate trained random forest regression model. Plots show the number of predicted rings on the x-axis with the ground-truth labels for each point on the y-axis and a dotted horizontal line representing perfect performance. At each SAGE score threshold, the regression error is calculated as the root mean-squared error (RMSE) and shown next to the threshold value.

# Supplement

Supplementary Figure 1

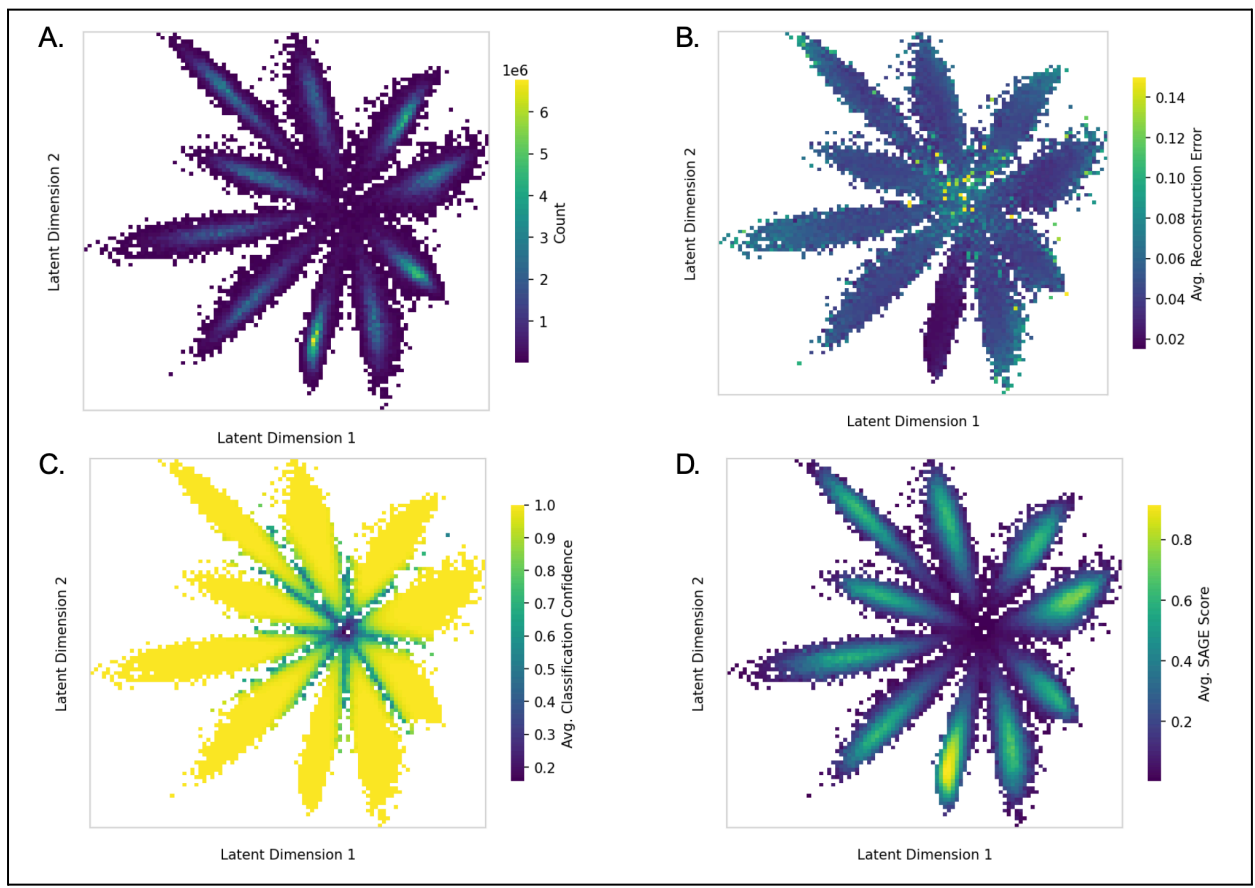

Binned output measures of the trained MNIST supervised autoencoder model. 2D latent embedding coordinates were determined for each training set image and the coordinate grid was divided into bins, 100 for each dimension ($n$ = 10,000). A) Overlay of train point density in the latent space, where a higher total number of points within a given bin is denoted with a lighter yellow color. Images primarily cluster into a dense asterisk pattern, where each finger of the asterisk represents a distinct class identity. B) Average MSE loss of reconstruction for all points in a given bin. Higher reconstruction error is denoted by a lighter yellow color, with the highest errors observed in bins close to the center of the embedding pattern or on the edges of clusters. C) Average classifier confidence is overlaid on the binned latent space. The heatmap distinguishes local areas of high confidence from low confidence bins. Darker, low confidence bins are concentrated in the center of the embedding pattern and in between embedded clusters. D) The average SAGE score per bin is shown using the same color heatmap. High-scoring bins are found towards the middle of each class cluster. SAGE scores decrease as bins extend to the outer edges of class clusters or towards the center of the latent space.

Supplementary Figure 2

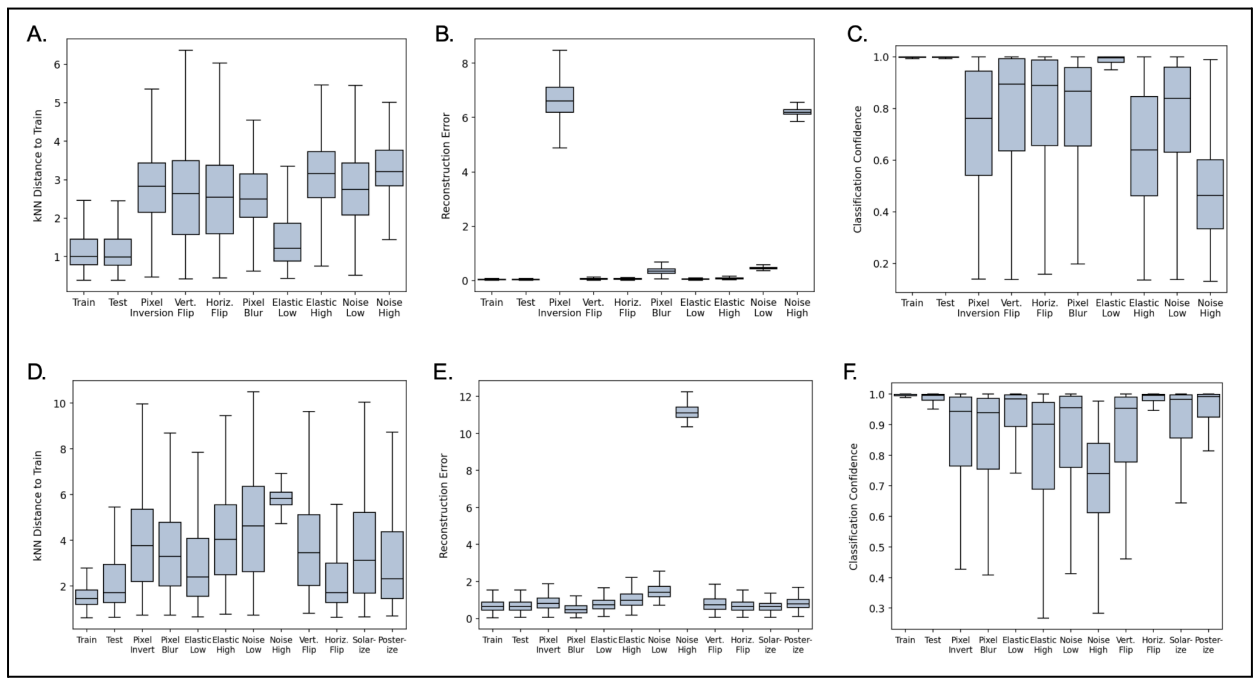

Boxplots showing trained supervised autoencoder output measures for MNIST (A - C) and CIFAR-10 (D - F) images. All models were trained on the train sets only. A, D) Boxplots show the average distance to the nearest training points ($k$ = 100) embedded in the 2D latent space. The y-axis shows the average kNN distance and the train, test and transformed test datasets are listed by name across the x-axis. B, E) The y-axis displays boxplots of the decoder's reconstruction error for each image as assessed by MSE loss, with the x-axis showing train, test and transformed test dataset names. C, F) Boxplots displaying the classification confidence of each image. Confidence is shown on the y-axis with the x-axis listing the train, test and transformed test datasets.

Supplementary Figure 3

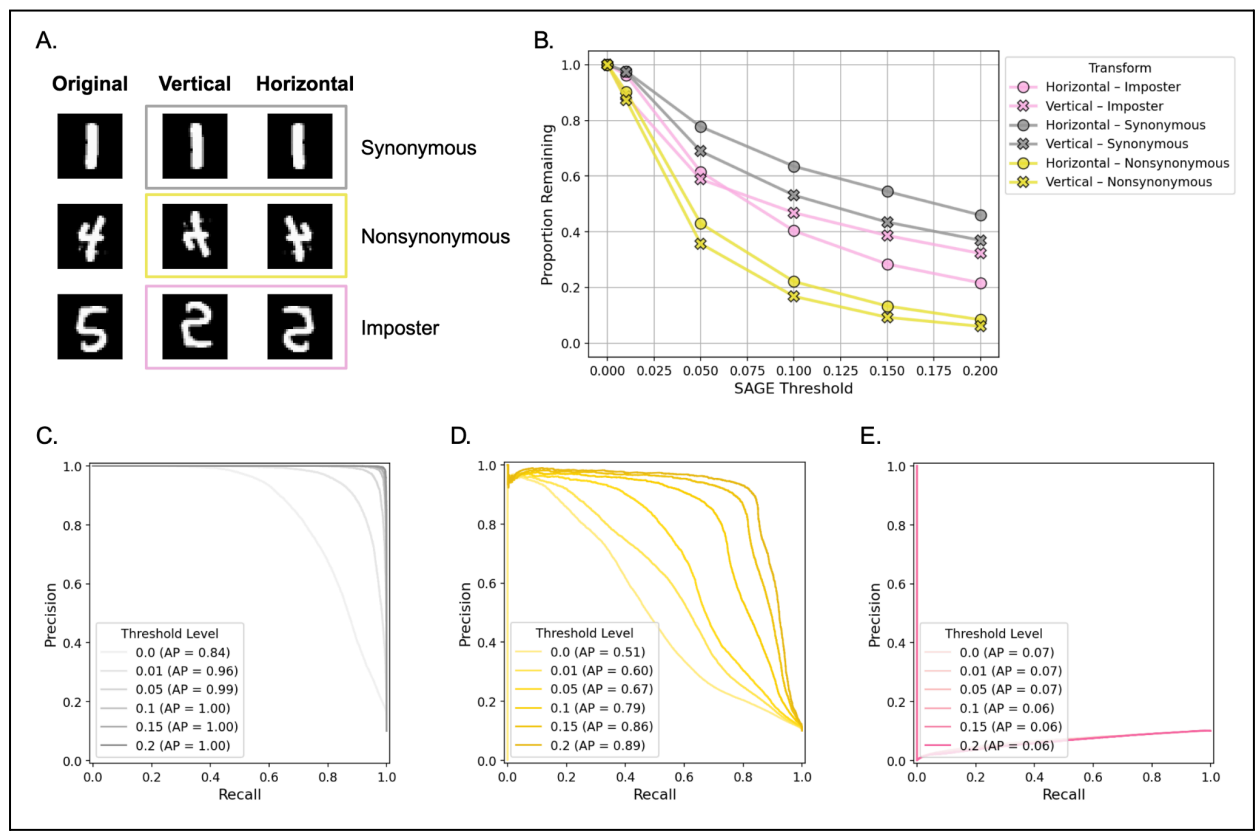

The effects of geometric flipping on SAGE score calculation and performance evaluation of a separate ResNet18 model for MNIST classification. A) Three examples of MNIST hand drawn digits are shown in the 'Original' column, with 'Vertical' and 'Horizontal' columns showing geometric transformations of the same images. Changes to legibility are categorized in three groups, with the name listed for each row. 'Synonymous' flips ('0', '1', '3' vertical, '8') are highlighted in grey and denote no change to the legibility of the image. 'Nonsynonymous' flips ('3' horizontal, '4', '6', '7', '9') are highlighted in yellow and change the digit to an unintelligible written pattern. 'Imposter' flips ('2', '5') are shown in pink and denote images that effectively become re-labeled, as depicted here where an image of a '5' is read as '2' when flipped vertically and horizontally. B) The trained MNIST model is used to calculate SAGE scores for each of the three flip categories, and the proportion of the original images remaining after 5 threshold values is plotted as a different color for each category and as a different marker to differentiate between horizontal and vertical transformation. C - E) At each threshold level, a separate ResNet18 model is used to calculate precision recall curves for the synonymous, nonsynonymous and imposter flip categories. Average precision values are shown in the plot legends beside the SAGE score threshold value.

Supplementary Figure 4

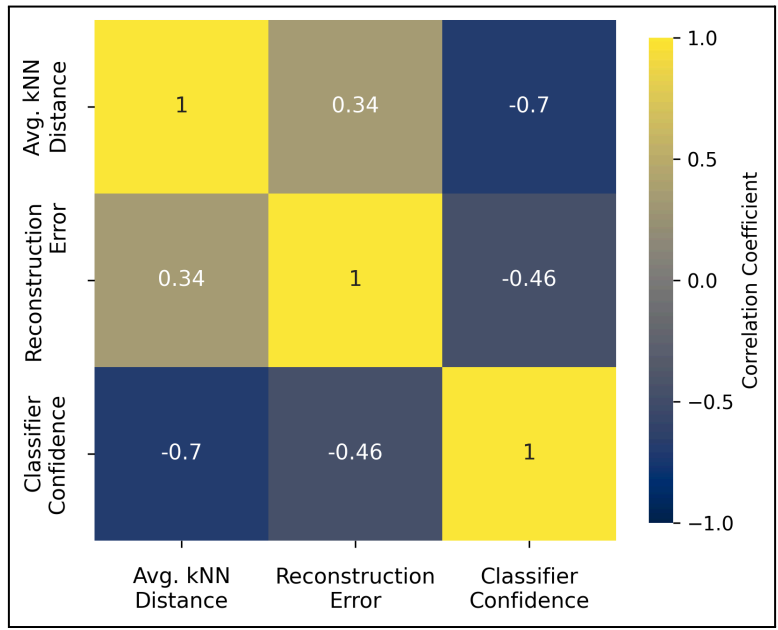

Matrix showing the Pearson correlation coefficient values for trained MNIST model output measures. The possible range extends from [-1.0 - 1.0]: perfect correlation is shown as a yellow box, with higher negative values shown in dark blue. Average kNN distance, reconstruction error and classifier confidence are repeated on both the x and y-axis ticks, so the correlation values for each distinct pair of measures is shown twice.

Supplementary Figure 5

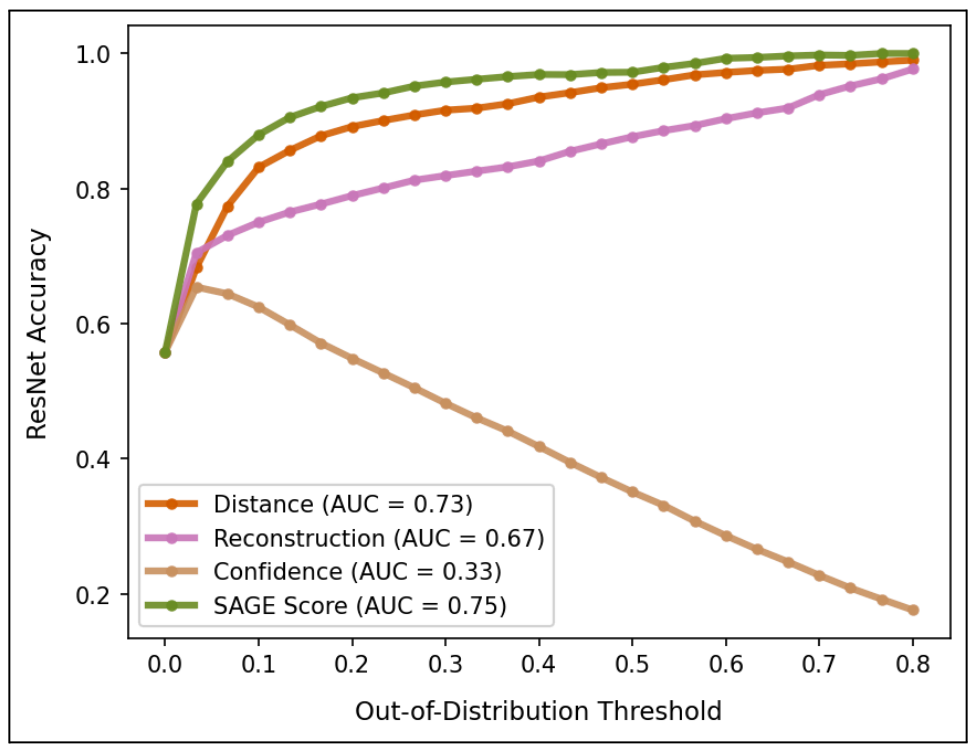

Changes in independent ResNet18 model accuracy when filtering transformed test images on individual or ensemble out-of-distribution threshold values. Threshold values are displayed on the x-axis while the separate model accuracy is shown on the y-axis. The accuracy at each threshold value is displayed as a marker for individual probability metrics (kNN distance, reconstruction error and classifier confidence) as well as the ensemble SAGE score. Markers are connected to form a curve for each out-of-distribution metric, and area under the curve (AUC) is shown alongside metric names in the figure legend. We include one additional transformation of MNIST lacking pixel value normalization to demonstrate classifier overconfidence.

Supplementary Figure 6

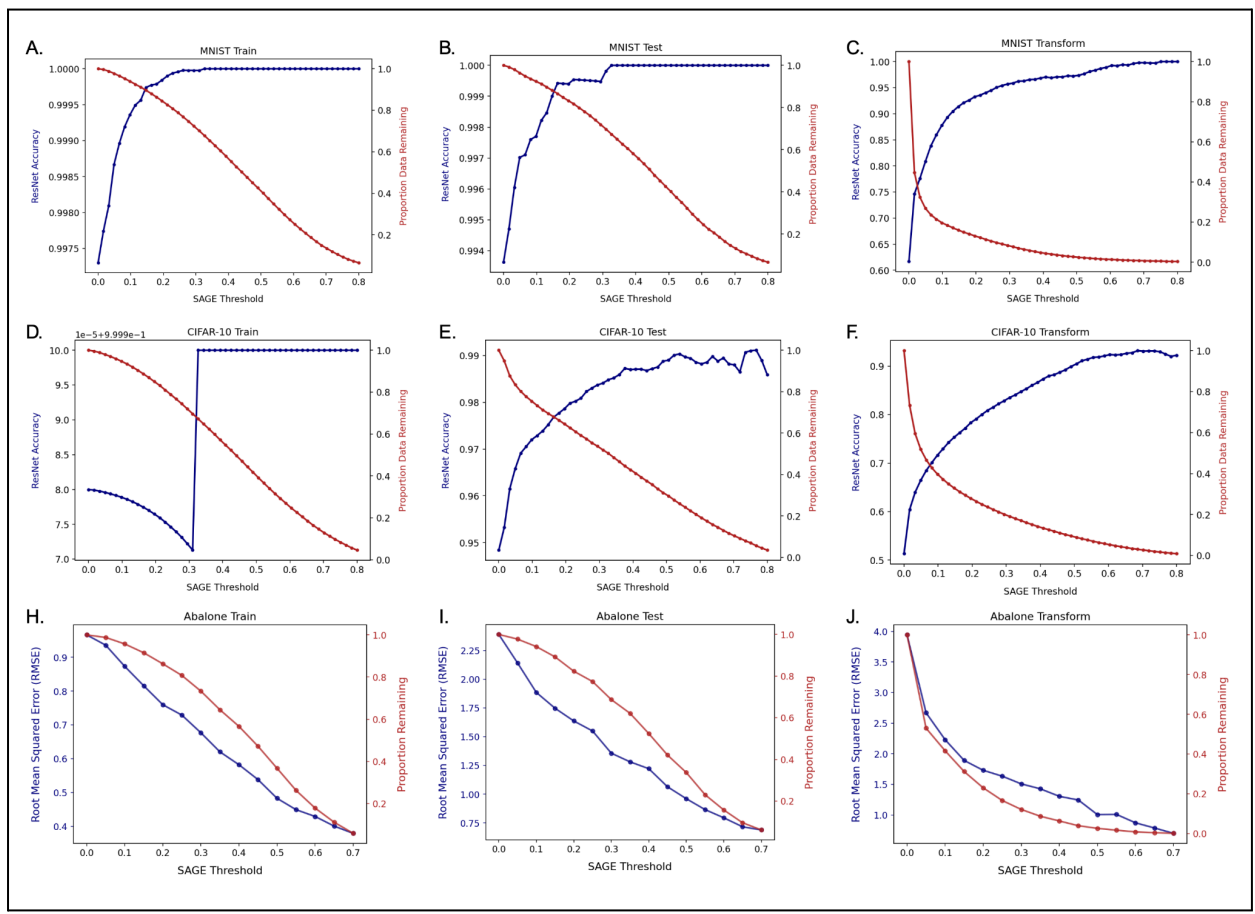

Extended performance analysis of SAGE score thresholding with a separate model for MNIST, CIFAR-10 and UCI Abalone data. For each plot, the proportion of the original dataset remaining after removing all points at or below the x-axis value is shown in dark red, with the proportion shown on the right hand y-axis. The left hand y-axis shows the performance of the separate prediction model at each threshold level, following the dark blue line. Y-axis scales vary between plots to allow for visualization of the performance trend line. A - C) MNIST train, test and transformed test SAGE score thresholds (range [0.0 - 0.8]), where performance is assessed with a separate ResNet18 classification accuracy. D - F) CIFAR-10 train, test and transformed test SAGE score thresholds (range [0.0 - 0.8]) with a separate ResNet34 model's predictive accuracy. H - J) UCI Abalone train, test and transformed test datasets after predictions with a separate random forest model. The left hand y-axis shows the root mean squared error (RMSE) of regression.

Supplementary Figure 7

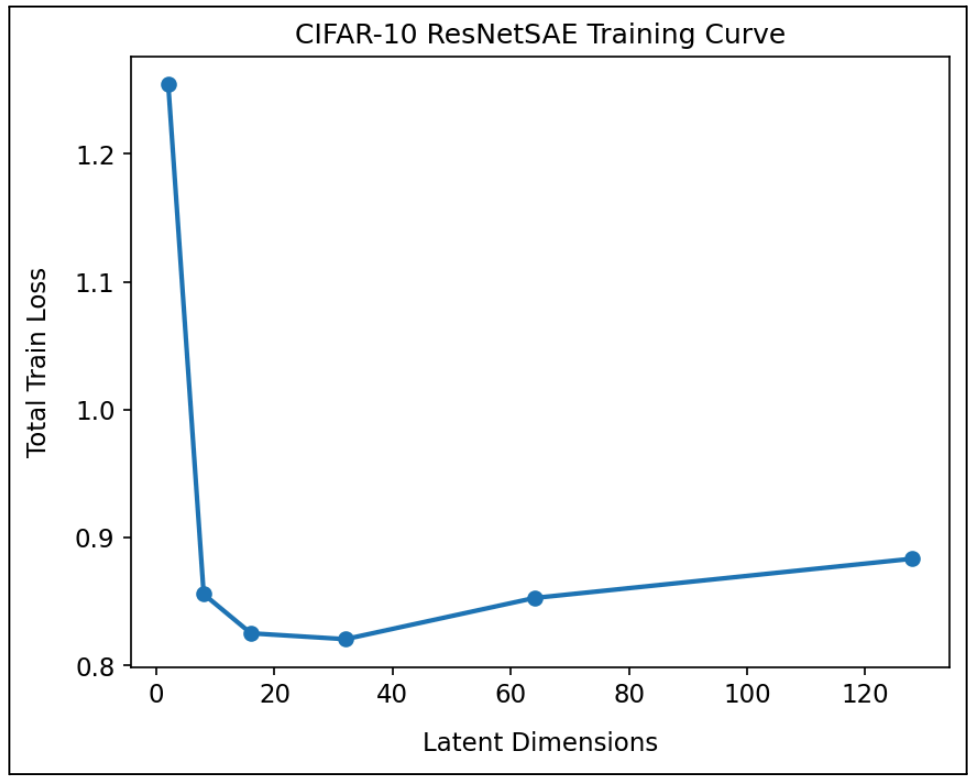

The lowest total training loss for ResNetSAE models using a variety of encoder output dimensionalities. The y-axis represents the CIFAR-10 training loss determined as the minimum loss over all training epochs. The x-axis shows the number of latent dimensions used as output of the encoder, and the input to the decoder and classifier modules (dimensions = [2, 8, 16, 32, 64, 128]). The total loss is the sum of the lowest loss values from training stages one and two.

Supplementary Figure 8

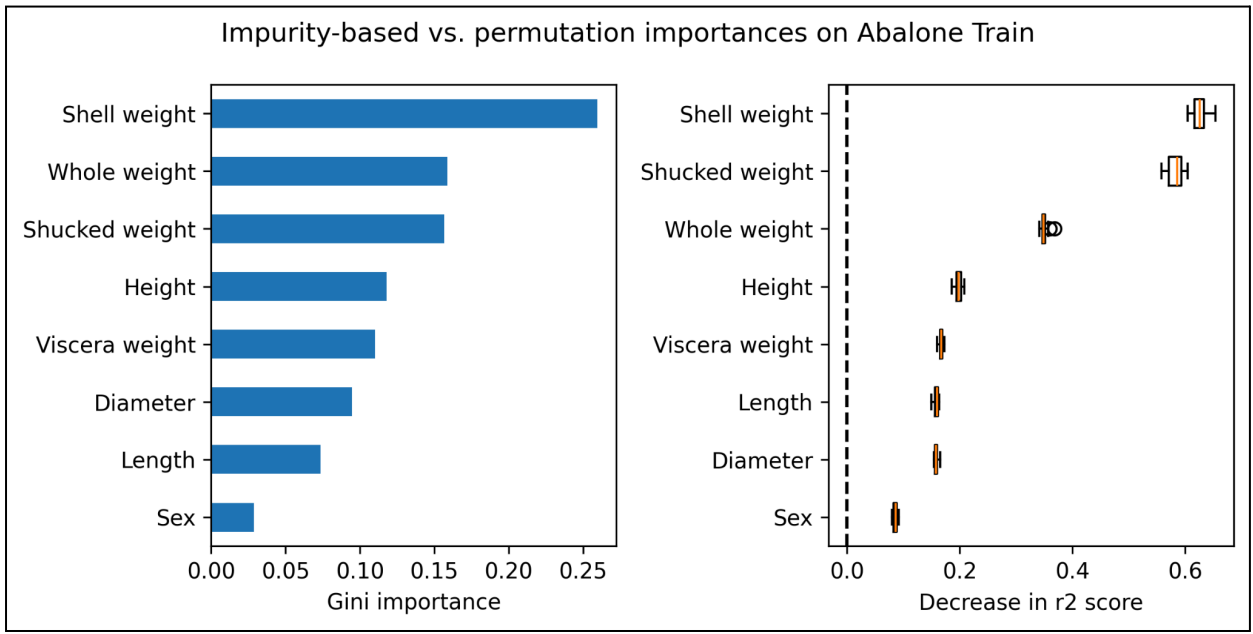

Feature importance of the separate random forest regressor model for the UCI Abalone dataset. The left panel shows impurity-based feature importance scores, assessed by Gini importance, and the right panel features the effect of removal of that feature on the overall r-squared score of predictions, with larger decreases in r-squared score (x-axis) equating to higher feature importance.

Supplementary Figure 9

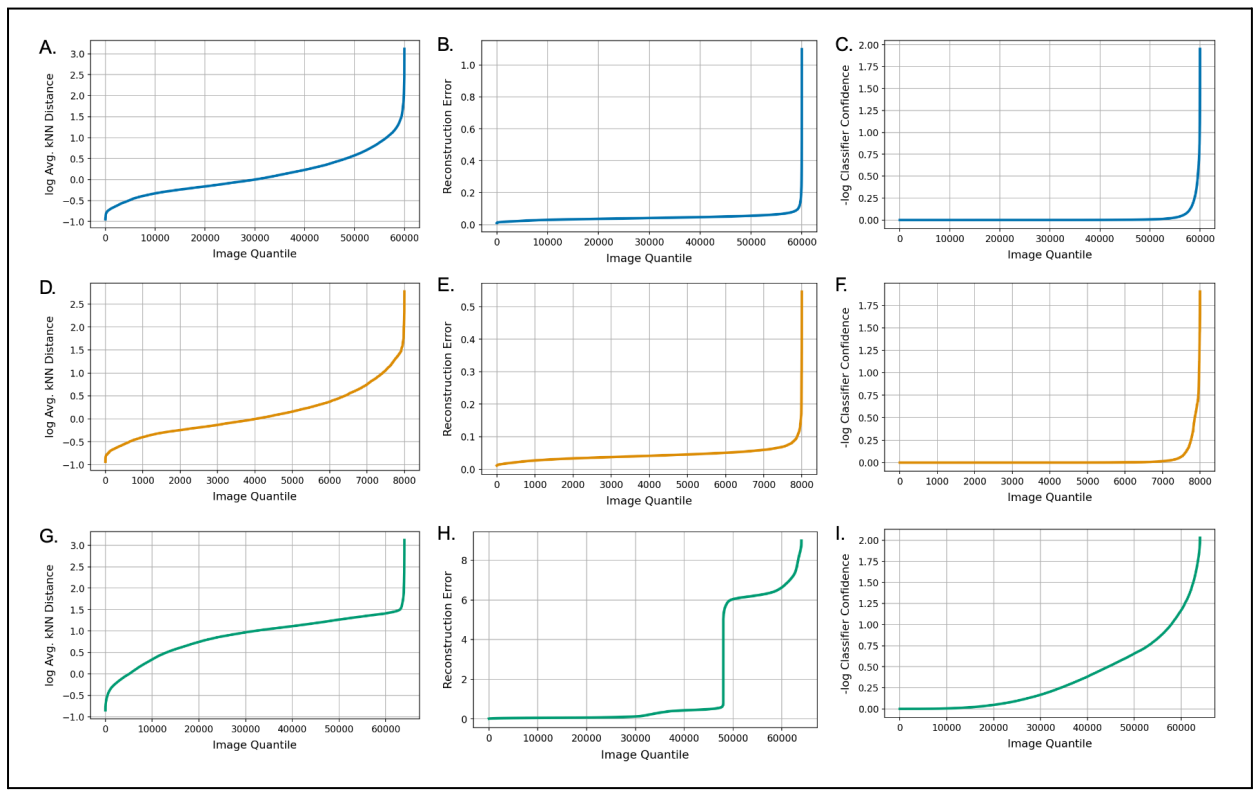

Quantile plots of the trained MNIST supervised autoencoder model outputs for train, test and transformed test sets. Log average kNN distance, reconstruction error and negative log classifier confidence make up the three y-axis measures from left to right, with the sorted measures for each dataset shown as quantiles on the x-axis. A - C) Blue curves for the training dataset ($n$ = 60,000) show the rate of change in kNN distance, reconstruction error and classifier confidence values. D - F) Orange curves represent the quantile plot for test set measures ($n$ = 8,000). G - I) Green curves are shown for the combined transformed dataset quantile plots ($n$ = 64,000).

Supplementary Figure 10

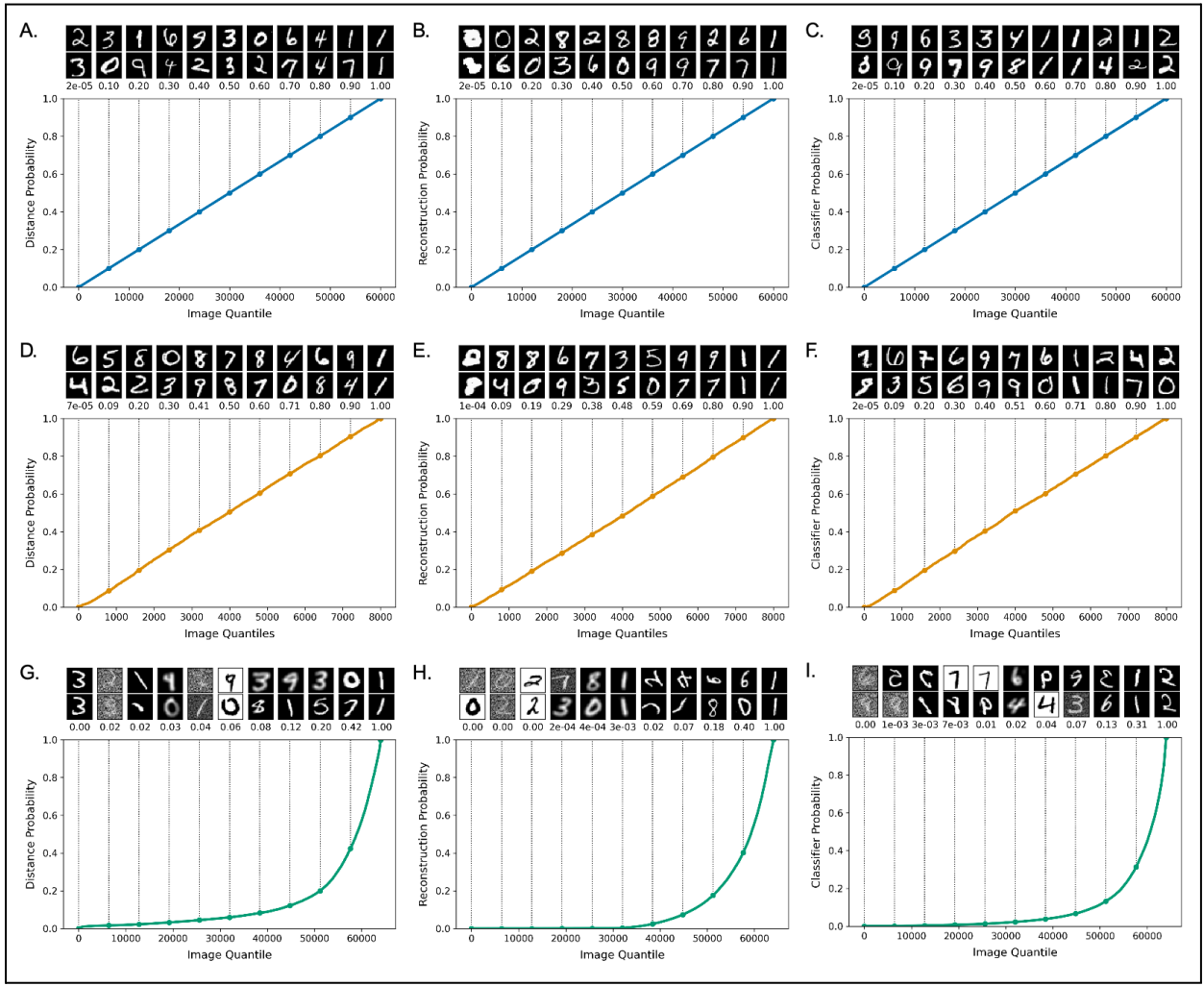

Quantile plots showing the individual exceedance probability values for each of the three output measures: kNN distance, reconstruction error and classification confidence (left to right). Probability values are sorted from low to high, with the y-axis showing one of the measure probabilities (range: [0.0 - 1.0]) and the x-axis denoting the sorted quantile. A vertical line leads to the score of the images at each decile, with the closest two images shown above the probability score value. A - C) Blue quantile curves for the train dataset kNN distance, reconstruction error and classification confidence. Curves for all train set probabilities represent a straight line, as probability values are incremented by the same amount for each training quantile. D - F) Test quantile curves for measure probabilities are shown in orange. G - I) Transformed test set probability quantile curves are shown in green.